%% file: main.tex
\PassOptionsToPackage{final}{graphicx}
\documentclass[times,preprint,10pt]{elsarticle}

\usepackage{amssymb}
\usepackage{amsmath}
\usepackage[T1]{fontenc}
\usepackage{subcaption}

\usepackage{xcolor}
\newcommand{\muted}[1]{{\color{black!45}#1}}
\newcommand{\gapcell}[1]{\textbf{#1}}

\usepackage{url}
\usepackage{xurl}
\usepackage{tikz}
\usepackage{pgfplots}
\usepackage{pgfplotstable}
\pgfplotsset{compat=1.18}
\usepgfplotslibrary{groupplots,fillbetween}
\usepackage{xcolor}
\usetikzlibrary{patterns}

\definecolor{ap1}{RGB}{31,119,180}
\definecolor{ar1}{RGB}{10,60,120}

\usepackage{graphicx}  % for \rotatebox
\usepackage{booktabs}
\usepackage[table]{xcolor}
\usepackage{multirow}
\usepackage[
  top=4.3cm,
  right=4.8cm,
  bottom=4.3cm,
  left=4.8cm
]{geometry}
\definecolor{model1}{RGB}{31,119,180}
\definecolor{model2}{RGB}{255,127,14}
\definecolor{model3}{RGB}{44,160,44}
\definecolor{model4}{RGB}{214,39,40}

\definecolor{minvisColor}{RGB}{52,101,164}   % deep desaturated blue
\definecolor{ctvisColor}{RGB}{204,120,50}    % muted orange-brown
\definecolor{dvisColor}{RGB}{78,154,6}       % muted green
\definecolor{cavisColor}{RGB}{176,46,68}     % muted red

\journal{Pattern Recognition}

\begin{document}

\begin{frontmatter}



\title{\fontsize{14}{16}\selectfont Mind the Gap: Disentangling Performance Bottlenecks in Video Instance Segmentation}


\author[amirkabir]{\fontsize{8}{10}\selectfont Danial Hamdi}
\author[amirkabir]{\fontsize{8}{10}\selectfont Fardin Ayar}
\author[amirkabir]{\fontsize{8}{10}\selectfont Mahdi Javanmardi\corref{cor1}}
\ead{mjavan@aut.ac.ir}
\cortext[cor1]{Corresponding author.}

\affiliation[amirkabir]{
    organization={Computer Engineering Department, Amirkabir University of Technology (Tehran Polytechnic)},
    addressline={No. 350, Hafez Ave., Valiasr Sq.}, 
    city={Tehran},
    postcode={15875-4413}, 
    state={Tehran},
    country={Iran}
}

% \affiliation[tokyo]{
%     organization={Graduate School of Information Science and Technology, The University of Tokyo},
%     addressline={7-3-1 Hongo},
%     city={Bunkyo},
%     postcode={113-8654},
%     state={Tokyo},
%     country={Japan}
% }

%% Abstract
\begin{abstract}

\input{0_abstract}

\end{abstract}

%% Graphical abstract intentionally omitted to avoid a blank submission page.

%% Keywords

\end{frontmatter}

% \linenumbers



\section{Introduction}
\label{sec1}
\input{1_introduction}

% ========================================================

\section{Related Work}
\label{sec2}
\input{2_related}

% ========================================================

\section{Method}
\label{sec3}
\input{3_method}

% ========================================================

\section{Experiments}
\label{sec4}
\input{4_experiments}

% ========================================================

\section{Discussion}
\label{sec5}
\input{5_discussion}

% ========================================================

\section{Conclusion}
\label{sec6}
\input{6_conclusion}

% ========================================================

\appendix
\input{7_appendix}

% \section*{Acknowledgements}
%% Acknowledgements must appear directly before the reference list.

\bibliographystyle{elsarticle-num} 
\bibliography{references}

\end{document}

%% file: 0_abstract.tex
In Video Instance Segmentation (VIS), classification, segmentation, and tracking objectives are jointly evaluated, but their individual contributions to performance loss remain opaque. We introduce a diagnostic framework that formulates identity and class assignment as an Integer Linear Program (ILP), yielding a model-agnostic oracle that hierarchically isolates each error source. Applied to seven VIS methods spanning online and offline paradigms across YouTube-VIS 2019/2021 and a diagnostic subset of OVIS, our analysis reveals a consistent picture. Tracking instability is a critical bottleneck for online methods, with gaps exceeding 20 AP under heavy occlusion, and grows sharply with video length and instance density. While semantic classification contributes meaningfully on standard benchmarks, its impact becomes negligible where tracking fails most. Although stronger backbones substantially lift default scores, they leave AP tracking gaps largely intact, confirming that temporal fragility is algorithmic rather than purely representational. To complement the oracle, we introduce TrackLens, a visual tool that translates gap magnitude into observable, query-level failure modes. Together, these tools provide a systematic foundation for targeting VIS's core challenge: robust long-term temporal association.

%% file: 1_introduction.tex
Video Instance Segmentation (VIS) is a fundamental computer vision task that involves the simultaneous detection, segmentation, and tracking of object instances in video sequences~\cite{yang2019vis}. While detection localizes and classifies instances, and segmentation delineates their precise boundaries within each frame, tracking is responsible for maintaining instance identities over time. This ensures that an object is consistently recognized as it moves, interacts with other objects, or undergoes appearance changes.

VIS methods are commonly divided into offline and online paradigms: offline methods process full videos or long clips to produce spatiotemporal masks, achieving strong accuracy at high memory and computational cost~\cite{cheng2021mask2formervis, wang2020end, wu2021seqformer, hwang2021ifc, lin2021proposereduce, heo2022vita, heo2023genvis}, whereas online methods segment frames independently and then associate instances over time~\cite{huang2022minvis, cao2020sipmask, kim2024visage, ying2023ctvis}, offering greater efficiency and real-time suitability~\cite{yang2021crossvis, huang2022minvis} but remaining vulnerable to error propagation and tracking drift under occlusion or rapid motion. Recent progress in both paradigms has been driven by query-based architectures such as Mask2Former~\cite{cheng2022masked}, where learnable \textit{queries} predict object masks and classes while shifting tracking from pixel-level matching to query-embedding association. Online frameworks such as MinVIS~\cite{huang2022minvis}, IDOL~\cite{wu2022defense}, and CTVIS~\cite{ying2023ctvis} exploit this structure through embedding-based matching, while VISAGE~\cite{kim2024visage} and CAVIS~\cite{lee2024context} add appearance and contextual cues; decoupled frameworks such as DVIS~\cite{zhang2023dvis} and DVIS++~\cite{zhang2023dvisplus} further separate segmentation, tracking, and temporal refinement to better handle long-term dependencies.

Despite these advancements, a fundamental question remains: \textit{to what extent is VIS performance limited by each task objective?} The tight coupling of classification, segmentation, and tracking in standard evaluation metrics obscures the individual contribution of each component to performance degradation. Without a systematic diagnostic analysis, it is unclear whether continued progress requires better temporal association mechanisms, improved per-frame segmentation quality, or enhanced semantic understanding. While prior analyses have provided partial insights, they often rely on heuristics or are bound to specific architectures \cite{yang2019vis, wu2022defense, qi2022ovis}.

In this work, we introduce the first large-scale oracle study of VIS to rigorously quantify these performance gaps. Adopting a deconstructive approach, we first isolate tracking error by formulating the ideal tracking problem as a discrete optimization task. This yields optimal spatiotemporal tracks that maximize spatiotemporal overlap with ground truth instances subject to correct classification constraints. Second, we relax this semantic requirement to isolate classification error. Consequently, the remaining performance gap is exclusively attributable to segmentation mask quality. While these objectives are not fully independent in practice — richer query representations may simultaneously improve mask quality and tracking discriminability — this overlap is precisely what makes a principled decomposition valuable: it separates what can be recovered through better representations from what requires rethinking the association logic itself.

We apply this framework to a diverse set of core VIS architectures, covering both online and offline variants across major benchmarks: YouTube-VIS 2019 (YTVIS19) \cite{yang2019vis}, YouTube-VIS 2021 (YTVIS21) \cite{yang2021vis}, and a diagnostic subset of Occluded-VIS (OVIS) \cite{qi2022ovis}. Our analysis reveals a clear stratification of failure modes. Tracking instability is the critical bottleneck for online methods — particularly under occlusion, where gaps exceed 20 AP — growing sharply with video length and instance density. Semantic classification, by contrast, plays a meaningful but secondary role on standard benchmarks and becomes nearly negligible precisely where tracking fails most. Offline methods show substantially smaller tracking gaps, and decoupled architectures further validate that separating temporal refinement from frame-level segmentation yields measurably more stable identity assignment.

Finally, to complement our quantitative framework, we introduce \textbf{TrackLens}, a visual diagnostic tool. Modern VIS models generate hundreds of query predictions per frame, each carrying its own mask and class logits, yet only a handful survive as final output tracks. TrackLens exposes this full internal prediction space — how individual queries relate to one another, and where the gap between model output and oracle assignment opens up. Together, our analytic framework and diagnostic tool provide the community with a systematic foundation to diagnose architectural bottlenecks and guide future research toward robust long-term temporal association.

%% file: 2_related.tex
\subsection{Paradigms in Video Instance Segmentation}

VIS originated with \emph{tracking-by-detection} paradigms, exemplified by MaskTrack R-CNN~\cite{yang2019vis}, which extends image detectors~\cite{he2017maskrcnn} by associating frame-wise predictions via learned embeddings and heuristic cues, mirroring classic multi-object tracking (MOT) algorithms~\cite{wokje2017sort}. Subsequent efforts like SipMask~\cite{cao2020sipmask} and CrossVIS~\cite{yang2021crossvis} improved feature discriminability. A paradigm shift occurred with \emph{spatio-temporal transformers} that model video clips holistically: VisTR~\cite{wang2020end} treats VIS as a direct sequence prediction problem, while IFC~\cite{hwang2021ifc} and SeqFormer~\cite{wu2021seqformer} introduced memory tokens and query decomposition. However, their inherent latency has catalyzed the recent transition toward efficient, online frameworks~\cite{huang2022minvis}.

\subsection{Query-Based Video Instance Segmentation}

Modern VIS approaches predominantly rely on the query-based paradigm~\cite{carion2020end,zhu2020deformable}, where learnable vectors represent object instances. MinVIS~\cite{huang2022minvis} challenged the necessity of complex architectures by demonstrating that discriminative image queries from models like Mask2Former~\cite{cheng2022masked} can inherently track objects. However, susceptibility to complex video dynamics prompted research into robust features (e.g., VISAGE~\cite{kim2024visage}, CAVIS~\cite{lee2024context}, MDQE~\cite{li2023mdqe}), and bridging the short-clip training and long-video inference gap---a challenge noted in memory-centric video segmentation models~\cite{cheng2021stcn,cheng2022xmem}---through unified label assignment or memory integration (GenVIS~\cite{heo2023genvis}, CTVIS~\cite{ying2023ctvis}). Decoupled architectures have also emerged to address long-term temporal dependencies by decomposing VIS into independent steps~\cite{zhang2023dvis,zhang2023dvisplus,zhou2024dvisdaq}, alongside universal prompt-based models such as TarVIS ~\cite{athar2023tarvis}, UniVS~\cite{li2024univs}, and GLEE~\cite{wu2023general}.

\subsection{Error Analysis in Video Instance Segmentation}

Prior work on VIS error analysis has largely taken the form of new metrics, diagnostic toolboxes, or benchmark-specific oracle studies. While these efforts provide valuable insights into failure modes, they do not explicitly decompose overall performance into mathematically separable sources of error.

In multi-object tracking, metrics like HOTA~\cite{luiten2021hota} explicitly decompose unified scores into detection and association accuracy. For image and video recognition, toolboxes like TIDE~\cite{bolya2020tide} and TIVE~\cite{jia2023tive} quantify the AP impact of specific spatial and temporal errors (e.g., identity switches) by operating on final, post-processed output files. However, these frameworks are inherently \emph{retrospective}: they tally the mistakes a model has already committed using heuristic post-processing, rather than establishing the maximum performance its raw predictions could theoretically achieve.

Alternatively, benchmark studies rely on oracle substitutions. Yang et al.~\cite{yang2019vis} and OVIS~\cite{qi2022ovis} used \emph{image} and greedy \emph{identity} oracles to replace specific outputs with ground truth. IDOL~\cite{wu2022defense} introduced \emph{clip oracles} to enforce perfect local associations. Yet, because these oracles were tailored for specific architectural studies—relying on greedy matching or short temporal windows—they approximate but mathematically cannot guarantee a globally optimal track. Consequently, the absolute upper bound of a model's temporal matching and classification capability remains unquantified.

Crucially, all these prior methods bypass the model's internal class-logit structure. We introduce a model-agnostic ILP oracle that operates directly on raw predictions. By solving a global spatiotemporal assignment problem under explicit class-consistency constraints, it establishes a \emph{provably optimal} tracking upper bound. This yields the first precise quantification of the performance gap attributable solely to suboptimal temporal matching, cleanly decoupled from classification capacity and mask quality.

%% file: 3_method.tex
\begin{figure*}[t]
    \centering
    \includegraphics[width=\textwidth]{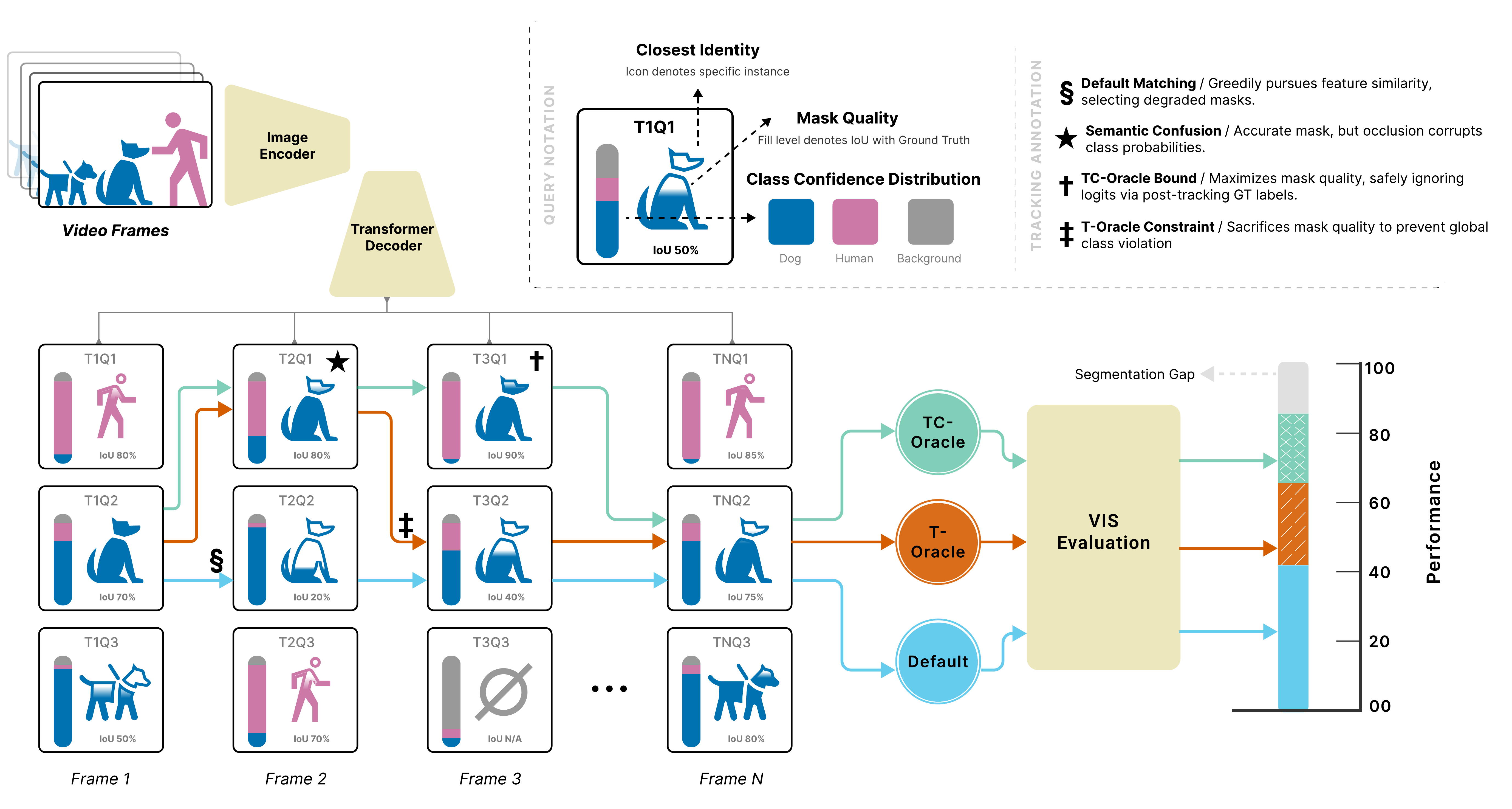}
    \caption{Overview of our hierarchical error decomposition framework. The figure illustrates the standard Video Instance Segmentation (VIS) pipeline alongside our proposed Tracking Oracle (T-Oracle) and Tracking+Classification Oracle (TC-Oracle), which systematically isolate tracking, classification, and mask quality errors.}
    \label{fig:method_main}
\end{figure*}

To systematically disentangle performance bottlenecks in VIS, we propose a hierarchical error decomposition framework isolating tracking, classification, and mask quality errors. As illustrated in Fig.~\ref{fig:method_main}, this framework visually and mathematically contrasts standard feature-based tracking against the theoretical bounds established by our TC-Oracle and T-Oracle. Section~\ref{sec:preliminary} formalizes the VIS task, demonstrating that optimal tracking is a discrete optimization problem. Section~\ref{sec:oracle} introduces our Integer Linear Programming oracle that hierarchically isolates these error sources. Section~\ref{sec:tracklens} presents TrackLens, a visualization tool for diagnosing the exposed failure modes.

% alt, longer version:

% To systematically disentangle the performance bottlenecks in VIS, we propose a hierarchical error decomposition framework that isolates tracking stability, semantic classification, and mask quality errors. We begin in Section~\ref{sec:preliminary} by formally defining the VIS task and demonstrating that optimal tracking is fundamentally a discrete optimization problem. This motivates Section~\ref{sec:oracle}, where we introduce an Integer Linear Programming (ILP) formulation—the Ideal Matching Oracle—that first establishes the tracking upper bound under realistic classification constraints, then relaxes semantic requirements to quantify classification headroom, leaving residual error attributable solely to mask quality. Finally, Section~\ref{sec:tracklens} describes TrackLens, a visualization tool designed to diagnose the specific failure modes exposed by this analysis.

\subsection{Preliminary}
\label{sec:preliminary}

\subsubsection{Problem Formulation}

% Formally, a video sequence \(V = \{I_t\}_{t=1}^{T}\) contains \(N\) ground-truth instances. Each instance \(i\) is defined by a category label \(c_i \in \mathcal{C}\) and a sequence of binary segmentation masks \(m_{i}^{(t)} \in \{0, 1\}^{H \times W}\). A VIS method produces a set of hypotheses, where each hypothesis \(j\) consists of a predicted category \(\tilde{c}_j\), a confidence score \(s_j\), and a predicted mask sequence \(\tilde{m}_{j}^{(t)}\).

Formally, a video sequence \(V = \{I_t\}_{t=1}^{T}\) of length \(T\) contains \(N\) ground-truth instances belonging to a category set \(\mathcal{C}\) of size \(K\). Each instance \(i\) is defined by a category label \(c_i \in \mathcal{C}\) and a sequence of binary segmentation masks \(m_{i}^{(t)} \in \{0, 1\}^{H \times W}\) for \(t=1,\dots,T\). A VIS method produces a set of hypotheses, where each hypothesis \(j\) consists of a predicted category \(\tilde{c}_j\), a confidence score \(s_j\), and a predicted mask sequence \(\tilde{m}_{j}^{(t)}\).

The standard AP and AR metrics assess hypotheses through two criteria: spatiotemporal overlap and a strict semantic constraint.

\textbf{1. Spatiotemporal IoU.} 
For a ground-truth instance \(i\) and a hypothesis \(j\), the intersection-over-union is computed over the entire video volume:
\begin{equation}
    \text{IoU}(i, j) = \frac{\sum_{t=1}^{T} \lvert m_{i}^{(t)} \cap \tilde{m}_{j}^{(t)} \rvert}{\sum_{t=1}^{T} \lvert m_{i}^{(t)} \cup \tilde{m}_{j}^{(t)} \rvert}
    \label{eq:st_iou}
\end{equation}
This metric couples mask quality with temporal consistency. Identity switches or fragmented tracks reduce the accumulated intersection while leaving the union large, penalizing even well-segmented per-frame predictions that fail to maintain coherent trajectories.

\textbf{2. Class Consistency.}
A hypothesis \(j\) matches instance \(i\) only if \(\tilde{c}_j = c_i\). Predictions with incorrect categories are rejected as false positives regardless of spatiotemporal overlap. 

Thus, high AP and AR require simultaneously achieving precise masks, stable identities, and correct semantic labels.

\subsubsection{Query-based Methods}

Query-based VIS methods generate, for each frame \(t\), a set of \(N_q\) queries, where each query consists of a predicted segmentation mask and a class logit vector over \(K\) categories. 
Standard inference then associates queries across frames to form tracks, ranks these tracks by confidence, and outputs the top-\(k\) predictions (typically \(k=10\) or \(k=25\)) for evaluation.

\subsection{Oracle Analysis}
\label{sec:oracle}

We begin by determining the \textit{tracking upper bound} via the \textbf{Tracking Oracle (T-Oracle)}: the optimal assignment of queries to ground-truth instances that maximizes spatiotemporal overlap while respecting the model's predicted class distributions. This is formulated as a discrete optimization problem. Solving this ILP isolates tracking error by showing the best achievable performance if the model had perfect temporal association, but still relied on its own semantic predictions.

We then construct the \textbf{Tracking+Classification Oracle (TC-Oracle)} by removing the class-consistency constraint entirely and assigning ground-truth labels to the T-Oracle's optimized tracks. This isolates classification error: the gap between T-Oracle and TC-Oracle quantifies how much performance is lost due to imperfect classification, while the remaining gap between TC-Oracle and perfect performance is attributable solely to mask quality.

\subsubsection{Formulation: The Tracking Oracle}

We first present the complete ILP formulation for the T-Oracle, consisting of three components: a maximization objective, class consistency constraints, and assignment constraints.

\textbf{Maximization Objective.} Directly maximizing the standard spatiotemporal IoU (Eq. \ref{eq:st_iou}) is a fractional programming problem that is computationally intractable for global optimization. To make this solvable via ILP, we approximate the global objective by maximizing the sum of frame-level IoUs. This linear proxy serves as a highly effective approximation to the volumetric IoU ratio.

We define a binary decision variable $x_{m,t,n}$ to represent the assignment of ground truth object $m$ to predicted query $n$ at frame $t$:
\begin{equation}
    x_{m,t,n} = \begin{cases} 
      1 & \text{if query } n \text{ at frame } t \text{ is assigned to object } m \\
      0 & \text{otherwise}
   \end{cases}
\end{equation}

The objective function maximizes the total intersection quality across the entire video sequence:
\begin{equation}
    \text{maximize} \sum_{m=1}^{M}\sum_{t=1}^{T}\sum_{n=1}^{N_q} \text{IoU}_{\text{frame}}(m,t,n) \cdot x_{m,t,n}
\end{equation}
where $\text{IoU}_{\text{frame}}(m,t,n)$ denotes the standard 2D IoU between the mask of ground truth object $m$ and query $n$ at frame $t$.

\textbf{Class Consistency Constraints.} In standard VIS, a track does not strictly require the highest score for its correct class; it merely needs a score high enough to survive the global top-\(k\) ranking. However, explicitly modeling this sorting logic within an ILP is computationally prohibitive. We therefore simplify this requirement by enforcing a stricter \textit{argmax consistency}: for a constructed track to be valid, the aggregated score for the ground-truth class must exceed that of any other class. We analyze the sensitivity of this constraint and its effect on feasibility rates in \ref{sec:app_argmax}. This ensures the oracle only generates tracks that would be confidently classified—and thus likely retained—by the model.

Assuming track-level class scores are aggregated by averaging frame-level logits—a prevalent strategy \cite{huang2022minvis, kim2024visage, ying2023ctvis, lee2024context, zhang2023dvis, zhang2023dvisplus}—let $L_{t,n,c}$ be the logit for class $c$ predicted by query $n$ at frame $t$. The average logit for a constructed track $m$ regarding class $c$ is:

\begin{equation}
    \bar{L}_{m,c} = \frac{1}{T} \sum_{t=1}^{T} \sum_{n=1}^{N_q} L_{t,n,c} \cdot x_{m,t,n}
\end{equation}
We enforce that for the ground truth class $c^*_m$, the score must strictly exceed all other classes $c'$ by a small margin $\epsilon$:
\begin{equation}
    \forall m, \forall c' \neq c^*_m: \quad \bar{L}_{m,c^*_m} \ge \bar{L}_{m,c'} + \epsilon
    \label{eq:class_consistency}
\end{equation}

\textbf{Assignment Constraints.} Finally, the optimization is subject to strict one-to-one assignment constraints. Each ground truth object must be assigned exactly one query per frame, and each query can be used at most once per frame:
\begin{equation}
    \forall m, t: \sum_{n=1}^{N_q} x_{m,t,n} = 1, \quad \forall t, n: \sum_{m=1}^{M} x_{m,t,n} \le 1
    \label{eq:assignment}
\end{equation}

Solving this ILP yields the \textbf{T-Oracle} upper bound.

\subsubsection{Variant: The Tracking+Classification Oracle}

To construct the TC-Oracle, we solve a relaxed variant of the above formulation where the class consistency constraint (Eq.~\ref{eq:class_consistency}) are entirely removed. Instead, we directly assign ground-truth labels $c_m^*$ to each optimized track after solving for the spatiotemporal assignments. This isolates the performance headroom attributable to imperfect classification, leaving only mask quality as the residual error source.

\subsection{TrackLens}
\label{sec:tracklens}

% While our ILP analysis quantifies the magnitude of the tracking gap, it does not explain the underlying causes of failure. To address this, we introduce \textit{TrackLens}, an interactive diagnostic tool that visualizes the temporal association process. It operates on standard frame-level outputs—masks and logits—rendering the affinity matrix to expose the raw similarity scores that drive tracking decisions.

% TrackLens facilitates side-by-side comparisons of different matching strategies, allowing researchers to contrast a model's default inference against the optimal assignment generated by our T-Oracle. By enabling the inspection of individual prediction attributes alongside pairwise metric overlays (e.g., cosine similarity or IoU), the tool helps pinpoint the precise drivers of tracking errors.

% This qualitative insight complements the quantitative rigor of our analytic framework, providing a holistic view of performance bottlenecks. Further implementation details and usage scenarios are provided in Appendix \ref{sec:app_tracklens}.

\begin{figure}[hbt!]
    \centering
    % width=\linewidth makes it fit the column perfectly
    \includegraphics[width=\linewidth]{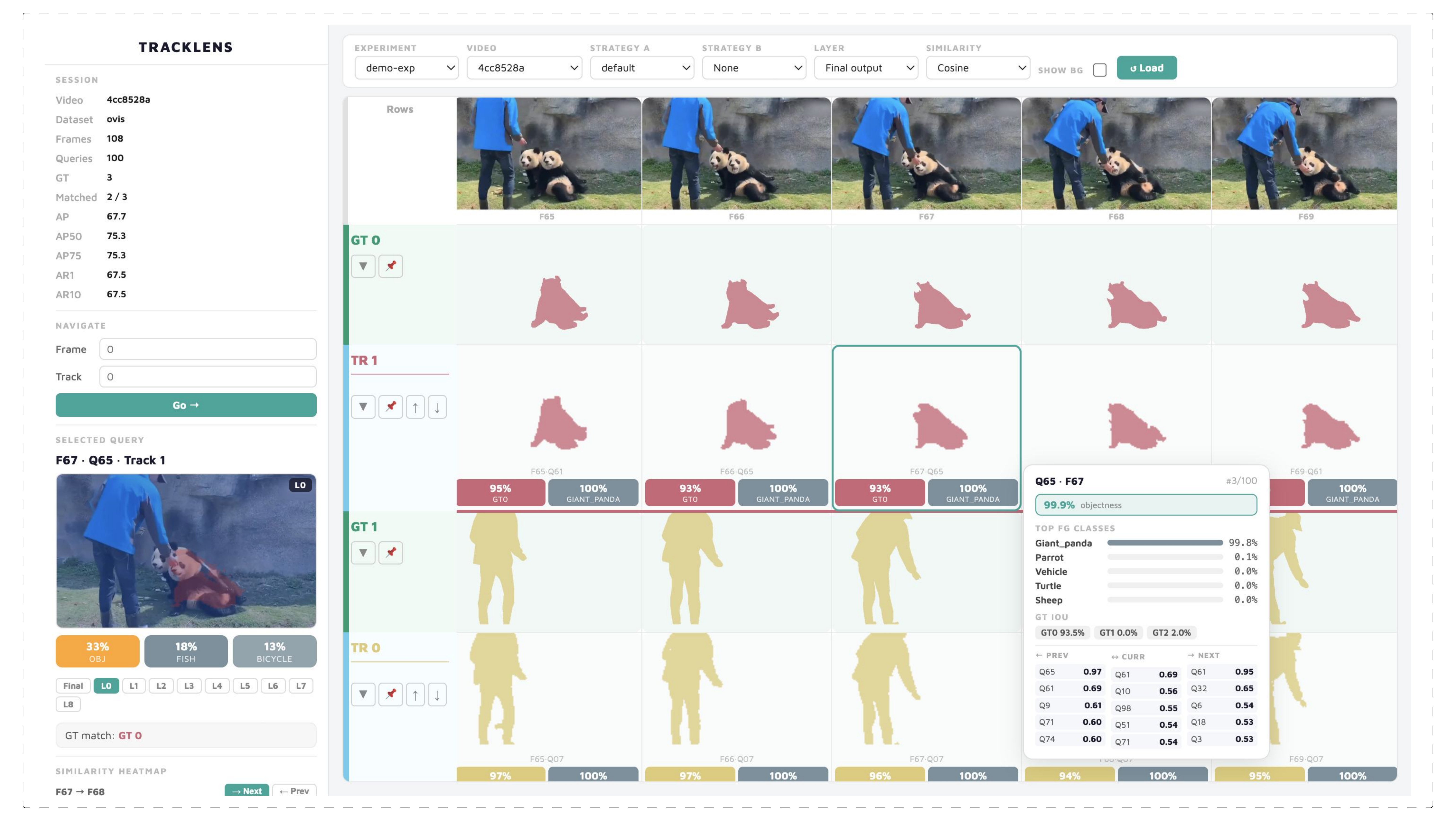}
    \caption{Overview of TrackLens. Rows represent query tracks across time, columns show video frames. The main panel displays the query-frame matrix, while the side panel shows relationship metrics. Top row: original video frames; white backgrounds: ground truth masks; colored backgrounds: strategy-specific tracklets.}
    \label{fig:tracklens}
\end{figure}

While our ILP analysis quantifies the tracking gap magnitude, it does not explain failure causes. We introduce TrackLens (Fig.~\ref{fig:tracklens}), an interactive diagnostic tool that visualizes frame-level outputs—masks and logits for all \(T \times Q\) queries—to expose how predictions evolve temporally and drive tracking decisions.

TrackLens enables side-by-side comparison of tracking decisions, contrasting a model's default inference against our T-Oracle's optimal assignment. By inspecting individual query attributes and computing pairwise metrics on demand (e.g., cosine similarity, IoU), researchers can pinpoint causes of tracking drift like identity switches or occlusion errors. 
Further details and usage scenario are provided in \ref{sec:app_tracklens}.

%% file: 4_experiments.tex
\subsection{Experimental Setup}

\paragraph{Datasets} We evaluate primarily on the YTVIS19 and YTVIS21 validation sets (302 and 421 videos across 40 categories)~\cite{yang2019vis,yang2021vis}. To analyze model behavior under heavy occlusion, we additionally use a diagnostic
subset of OVIS (25 categories)~\cite{qi2022ovis}. Since OVIS validation annotations are not publicly available, we sample 100 videos from the OVIS training set, stratified by video length and instance density to preserve the training distribution across both axes (Table~\ref{tab:ovis_stats}). This subset is used solely for diagnostic analysis; all main results are reported on YTVIS.

\input{4_experiments_plots_ovis_stats}

\paragraph{Baselines} We analyze a suite of state-of-the-art VIS methods: MinVIS \cite{huang2022minvis}, VISAGE \cite{kim2024visage}, CTVIS \cite{ying2023ctvis}, GenVIS \cite{heo2023genvis}, CAVIS \cite{lee2024context}, DVIS \cite{zhang2023dvis}, and DVIS++ \cite{zhang2023dvisplus}—including their offline variants where available. All baselines are evaluated using their official ResNet-50 \cite{he2015resnet} backbone weights. The only exception is CAVIS-Offline, which uses ViT-L due to the absence of publicly released R50 weights.

\paragraph{Evaluation Metrics} We follow the standard evaluation protocol, reporting Average Precision (AP) and Average Recall (AR). AP is averaged over 10 IoU thresholds from 0.50 to 0.95 (step 0.05), with $\text{AP}_{50}$ and $\text{AP}_{75}$ denoting fixed thresholds at 0.50 and 0.75. For recall, $\text{AR}_1$ and $\text{AR}_{10}$ measure average recall when evaluation is restricted to at most 1 and 10 predicted instances per video, respectively.

\paragraph{Implementation details} While the oracle formulation supports different temporal aggregation schemes for track classification (e.g., max pooling), we describe average pooling in the method section and match each model's strategy when enforcing class consistency during evaluation. The margin $\epsilon$ in Equation 5 is set to 0.001. 

% When reporting default performance, we use each method's standard inference but constrain each predicted track to a single label by ranking tracks using only their highest-scoring class.

We use OR-Tools \cite{perron2025ortools} as the ILP solver, with a per-video time limit of 180 seconds. When the ILP finds no feasible solution at the full set of ground-truth instances, we iteratively reduce the number of target instances and retain the assignment with the highest total IoU among feasible solutions. If no feasible assignment is found for any subset, we fall back to the model's original output for that video.

\subsection{Main Results}

\paragraph{Tracking bottleneck on YTVIS}
We first quantify the pure tracking deficit by comparing default inference against the Tracking Oracle on YTVIS19 and YTVIS21 (Tables~\ref{tab:yt_online} and~\ref{tab:yt_offline}). For online methods, tracking gaps span 5--17\,AP on YTVIS19 and 5--14\,AP on YTVIS21, revealing that temporal association is a major, architecture-dependent error source. Similarity-based methods suffer most: MinVIS, VISAGE, and CTVIS each lose 13--17\,AP on YTVIS19, indicating their errors stem largely from suboptimal query-to-instance matching rather than recognition or mask quality. Methods with richer temporal mechanisms show smaller gaps (CAVIS and DVIS lose 9--11\,AP), while GenVIS—whose memory-based assignment provides richer temporal context—exhibits the smallest tracking gap among online methods (5\,AP on both benchmarks).

\input{4_experiments_tables_yt_online}

Among offline methods, the picture diverges sharply. DVIS offline nearly closes the tracking bottleneck on standard benchmarks: its tracking gap is only +0.6\,AP on YTVIS19 after decoupled temporal refinement. However, it carries the largest classification gap (+15.9\,AP on YTVIS19). Conversely, CAVIS offline—evaluated with a ViT-L backbone due to absent public R50 weights—presents the opposite profile: the highest default AP (66.4 on YTVIS19) alongside the smallest classification gap (4.1\,AP), indicating a more balanced error decomposition.

\input{4_experiments_tables_yt_offline}

\paragraph{Tracking bottleneck on OVIS}
This trend sharpens considerably on OVIS (Table~\ref{tab:ovis}), where heavy occlusion and crowded scenes expose the limits of all online methods. Tracking gaps jump to 12--23\,AP—nearly double those on YTVIS—with MinVIS and CTVIS losing roughly 23 and 19\,AP strictly to tracking failures. Meanwhile, AR$_1$ improves only modestly (1--4\,points), confirming the oracle primarily recovers identities for already-detected objects rather than surfacing new ones. Consequently, temporal association emerges as the most fragile component of online methods under extreme occlusion.

\input{4_experiments_tables_ovis}

\paragraph{Semantic classification}
On standard benchmarks, classification is a secondary but non-trivial bottleneck. The TC-Oracle adds 4--13\,AP over the T-Oracle on YTVIS, confirming that mask quality and localization remain the dominant residual errors. This picture changes markedly on OVIS, where classification gaps collapse below 1.5\,AP for most methods. This sharp contrast reveals that semantic recognition is surprisingly robust to heavy occlusion. The models typically predict the correct categories at the frame level but fail to associate them temporally; once the tracking oracle restores identity, explicit semantic correction yields negligible additional gains.

\begin{figure*}[t]
    \centering
    \input{4_experiments_plots_ytvis_scale}
    \caption{Tracking and classification gaps across backbone scales on YouTube-VIS 2021. For each method, the left and right bar pairs show Average Precision (AP) and Average Recall (AR\textsubscript{1}), comparing Base (ResNet-50) and Scaled (Swin-L, except ViT-L for DVIS++) architectures. Solid segments represent default inference performance, orange hatched segments indicate the tracking gap (T-Oracle\,$-$\,Default), and green crosshatched segments denote the classification gap (TC-Oracle\,$-$\,T-Oracle). While backbone scaling improves default performance, a persistent AP tracking gap remains across methods. In contrast, AR\textsubscript{1} exhibits much smaller tracking gaps, reflecting its lower sensitivity to temporal identity switches compared to the strict spatiotemporal IoU requirements of AP.}
    \label{fig:main-scale}
\end{figure*}
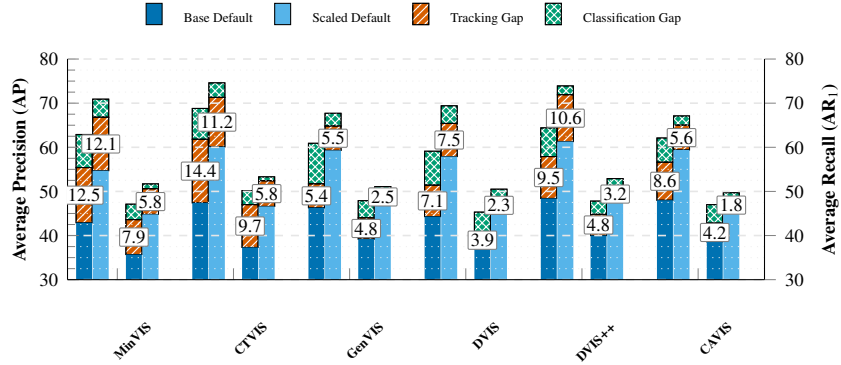

\paragraph{Does scaling solve tracking?}
A key question is whether the large tracking gaps reflect limited feature capacity. To test this, we repeat our oracle analysis with stronger backbones (Swin-L, except ViT-L for DVIS++; Fig.~\ref{fig:main-scale}). 
Scaling raises default AP by 11--14\,points on YTVIS21 and roughly halves classification gaps across all methods, confirming that richer representations substantially improve semantic recognition. Conversely, AP tracking gaps remain largely stable or even increase slightly for most methods (DVIS++: 9.5$\rightarrow$10.6, DVIS: 7.1$\rightarrow$7.5), with only modest reductions for others (CTVIS: 14.4$\rightarrow$11.2, CAVIS: 8.6$\rightarrow$5.6). Meanwhile, AR$_1$ tracking gaps consistently decrease (by 1.6--3.9\,points), indicating that while stronger representations improve object recall, they cannot repair the spatiotemporal consistency failures penalized by AP. 
The same conclusion holds on OVIS (Table~\ref{tab:ovis}): Swin-L raises default AP by 3--8 points across methods yet leaves tracking gaps in the 8--24\,AP range, reinforcing that representational scaling cannot compensate for structural matching failures under heavy occlusion.
Together, these results demonstrate that tracking fragility is primarily algorithmic rather than representational: the bottleneck lies in the association logic, not the visual features driving it.

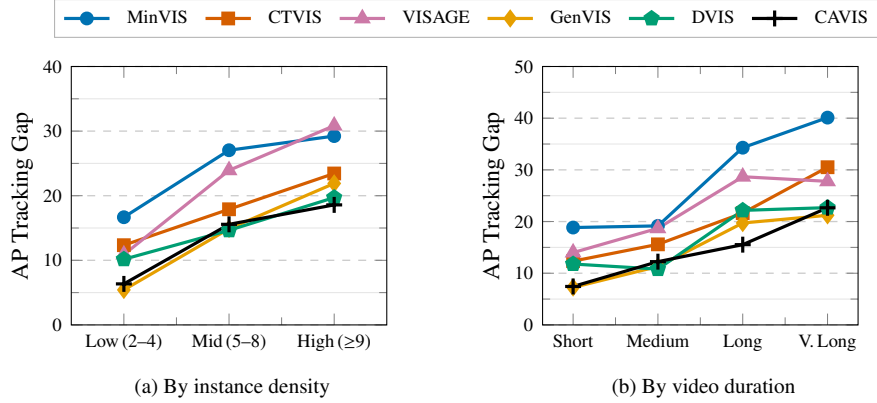
\begin{figure}[t]
\centering

% Colors — Okabe-Ito, consistent with Figure 2
\definecolor{colorMinVIS}{HTML}{0072B2}
\definecolor{colorCTVIS}{HTML}{D55E00}
\definecolor{colorGenVIS}{HTML}{E69F00}
\definecolor{colorDVIS}{HTML}{009E73}
\definecolor{colorVISAGE}{HTML}{CC79A7}
\definecolor{colorCAVIS}{HTML}{000000}

% Shared legend exported from the density axis (see that file)
\ref{legendOracleGap}\\[4pt]

\begin{subfigure}[b]{0.48\linewidth}
    \centering
    \input{4_experiments_plots_density}
    \caption{By instance density}
    \label{fig:oracle_gap_density}
\end{subfigure}
\hfill
\begin{subfigure}[b]{0.48\linewidth}
    \centering
    \input{4_experiments_plots_length}
    \caption{By video duration}
    \label{fig:oracle_gap_duration}
\end{subfigure}

\caption{
    AP tracking gap by instance density and video duration on the
    OVIS diagnostic split (online methods, ResNet-50). Tracking gaps grow
    consistently with both complexity dimensions, with the sharpest
    increases in dense (${\geq}9$ instances) and long (${\geq}90$ frame)
    videos. Bin sample sizes: density: $n{=}48/33/19$; duration: $n{=}35/45/13/7$.
}
\label{fig:oracle_gap_combined}
\end{figure}

\paragraph{Complexity analysis: length and density}
Finally, we examine how tracking error varies with video complexity by grouping OVIS videos by instance density and sequence length
(Fig.~\ref{fig:oracle_gap_combined}).
The trend is steep and consistent: for all online methods, tracking gaps grow sharply as videos become longer and more crowded.
MinVIS's gap grows from 17\,AP at low instance density to 29\,AP at high density; VISAGE follows an even steeper trajectory, growing from
11 to 31\,AP.
Even the most stable methods show meaningful degradation across the complexity spectrum.
These patterns confirm that current online designs lack mechanisms for maintaining identity over extended temporal horizons—a structural limitation that backbone scaling, as shown above, cannot address.

%% file: 4_experiments_plots_ovis_stats.tex
\begin{table}[htb]
    \centering
    \caption{Distribution of the OVIS diagnostic subset across video length and instance density bins (N=100).}
    \label{tab:ovis_stats}
    \begin{tabular}{llc}
    \hline
    \textbf{Characteristic} & \textbf{Category} & \textbf{Count} \\
    \hline
    \textbf{Video Length} & Short (0--44 frames) & 35 \\
     & Medium (45--89 frames) & 45 \\
     & Long (90--149 frames) & 13 \\
     & Very Long ($\ge$ 150 frames) & 7 \\
    \hline
    \textbf{Instance Density} & Low (2--4 instances) & 48 \\
     & Mid (5--8 instances) & 33 \\
     & High (9--12 instances) & 14 \\
     & Very High ($\ge$ 13 instances) & 5 \\
    \hline
    \end{tabular}
\end{table}

%% file: 4_experiments_tables_yt_online.tex
% Preamble (unchanged from before):
% \usepackage{xcolor}
% \newcommand{\muted}[1]{{\color{black!45}#1}}
% \newcommand{\gapcell}[1]{\textbf{#1}}

\begingroup
\setlength{\tabcolsep}{2pt} % submission

\begin{table*}[htbp]
\centering
\caption{Default (Def.), T-Oracle (TO), and TC-Oracle (TCO) results for
\textbf{online} VIS methods on YouTube-VIS 2019 and 2021~\cite{yang2019vis, yang2021vis}.
TO isolates tracking error; TCO additionally removes classification
inconsistencies. $\Delta$AP and $\Delta$AR$_1$ are relative to the
Default baseline.}
\label{tab:yt_online}

\resizebox{\textwidth}{!}{%

\begin{tabular}{l l @{\hskip 2pt} ccccccc @{\hskip 4pt} ccccccc}
  \toprule
    \multirow{2}{*}{Method}
    & \multirow{2}{*}{Var.}
    & \multicolumn{7}{c}{YTVIS19}
    & \multicolumn{7}{c}{YTVIS21} \\
  \cmidrule(lr){3-9} \cmidrule(lr){10-16}
    &
    & AP & \muted{AP$_{50}$} & \muted{AP$_{75}$} & AR$_1$ & \muted{AR$_{10}$} & $\Delta$AP & $\Delta$AR$_1$
    & AP & \muted{AP$_{50}$} & \muted{AP$_{75}$} & AR$_1$ & \muted{AR$_{10}$} & $\Delta$AP & $\Delta$AR$_1$ \\
  \midrule

    \multirow{3}{*}{MinVIS~\cite{huang2022minvis}}
      & Def. & 45.3 & \muted{66.7} & \muted{49.9} & 42.6 & \muted{50.7} & --   & --
             & 42.9 & \muted{64.5} & \muted{46.8} & 35.7 & \muted{48.5} & --   & --  \\
      & TO   & 62.0 & \muted{87.1} & \muted{68.1} & 52.6 & \muted{67.2} & \gapcell{+16.7} & +10.0
             & 55.4 & \muted{78.9} & \muted{62.7} & 43.6 & \muted{63.7} & \gapcell{+12.5} & +7.9  \\
      & TCO  & 69.8 & \muted{96.9} & \muted{77.0} & 57.8 & \muted{73.3} & \gapcell{+24.5} & +15.2
             & 62.9 & \muted{92.2} & \muted{69.9} & 47.1 & \muted{68.5} & \gapcell{+20.0} & +11.4 \\

  \cmidrule(lr){1-16}

    \multirow{3}{*}{VISAGE~\cite{kim2024visage}}
      & Def. & 52.2 & \muted{74.6} & \muted{58.0} & 46.4 & \muted{57.0} & --   & --
             & 49.5 & \muted{71.1} & \muted{54.1} & 40.3 & \muted{53.7} & --   & --  \\
      & TO   & 65.9 & \muted{87.6} & \muted{73.9} & 56.9 & \muted{72.8} & \gapcell{+13.7} & +10.4
             & 58.8 & \muted{81.6} & \muted{65.2} & 46.5 & \muted{67.8} & \gapcell{+9.2}  & +6.2  \\
      & TCO  & 72.9 & \muted{97.1} & \muted{81.2} & 59.6 & \muted{77.0} & \gapcell{+20.7} & +13.1
             & 66.2 & \muted{92.0} & \muted{73.6} & 50.2 & \muted{71.9} & \gapcell{+16.7} & +9.8  \\

  \cmidrule(lr){1-16}

    \multirow{3}{*}{GenVIS~\cite{heo2023genvis}}
      & Def. & 48.3 & \muted{67.6} & \muted{53.3} & 43.8 & \muted{53.4} & --   & --
             & 46.3 & \muted{66.3} & \muted{50.8} & 39.2 & \muted{51.6} & --   & --  \\
      & TO   & 53.3 & \muted{72.8} & \muted{59.5} & 49.8 & \muted{63.1} & \gapcell{+5.0}  & +6.1
             & 51.7 & \muted{71.7} & \muted{56.8} & 44.0 & \muted{60.4} & \gapcell{+5.4}  & +4.8  \\
      & TCO  & 66.0 & \muted{91.8} & \muted{72.0} & 56.0 & \muted{70.8} & \gapcell{+17.7} & +12.2
             & 60.9 & \muted{85.9} & \muted{66.4} & 47.9 & \muted{65.6} & \gapcell{+14.6} & +8.7  \\

  \cmidrule(lr){1-16}

    \multirow{3}{*}{CTVIS~\cite{ying2023ctvis}}
      & Def. & 51.3 & \muted{73.5} & \muted{55.7} & 43.9 & \muted{54.9} & --   & --
             & 47.4 & \muted{68.9} & \muted{51.4} & 37.3 & \muted{52.1} & --   & --  \\
      & TO   & 64.3 & \muted{87.8} & \muted{70.3} & 53.6 & \muted{69.2} & \gapcell{+13.0} & +9.7
             & 61.8 & \muted{83.7} & \muted{71.5} & 47.0 & \muted{66.9} & \gapcell{+14.4} & +9.7  \\
      & TCO  & 73.5 & \muted{97.7} & \muted{82.1} & 59.6 & \muted{76.7} & \gapcell{+22.2} & +15.7
             & 68.8 & \muted{94.6} & \muted{77.0} & 50.2 & \muted{72.2} & \gapcell{+21.4} & +12.8 \\

  \cmidrule(lr){1-16}

    \multirow{3}{*}{DVIS~\cite{zhang2023dvis}}
      & Def. & 47.8 & \muted{69.4} & \muted{53.3} & 42.8 & \muted{53.9} & --   & --
             & 44.3 & \muted{64.7} & \muted{48.4} & 37.0 & \muted{51.5} & --   & --  \\
      & TO   & 57.7 & \muted{80.9} & \muted{62.7} & 50.0 & \muted{62.3} & \gapcell{+9.9}  & +7.2
             & 51.4 & \muted{74.0} & \muted{56.4} & 40.9 & \muted{57.1} & \gapcell{+7.1}  & +3.9  \\
      & TCO  & 65.9 & \muted{94.0} & \muted{71.5} & 55.5 & \muted{69.0} & \gapcell{+18.1} & +12.7
             & 59.1 & \muted{85.7} & \muted{64.9} & 45.3 & \muted{62.5} & \gapcell{+14.8} & +8.3  \\

  \cmidrule(lr){1-16}

    \multirow{3}{*}{DVIS++~\cite{zhang2023dvisplus}}
      & Def. & 52.2 & \muted{75.6} & \muted{56.2} & 45.6 & \muted{56.9} & --   & --
             & 48.4 & \muted{69.7} & \muted{52.4} & 40.0 & \muted{55.2} & --   & --  \\
      & TO   & 59.4 & \muted{82.9} & \muted{64.8} & 49.6 & \muted{63.4} & \gapcell{+7.2}  & +4.0
             & 57.9 & \muted{82.9} & \muted{64.0} & 44.8 & \muted{63.5} & \gapcell{+9.5}  & +4.8  \\
      & TCO  & 67.1 & \muted{93.9} & \muted{72.9} & 56.3 & \muted{70.2} & \gapcell{+14.9} & +10.7
             & 64.4 & \muted{92.5} & \muted{71.0} & 47.8 & \muted{67.6} & \gapcell{+16.0} & +7.8  \\

  \cmidrule(lr){1-16}

    \multirow{3}{*}{CAVIS~\cite{lee2024context}}
      & Def. & 51.4 & \muted{71.9} & \muted{57.0} & 45.2 & \muted{55.8} & --   & --
             & 48.0 & \muted{70.1} & \muted{52.9} & 38.6 & \muted{53.2} & --   & --  \\
      & TO   & 62.5 & \muted{84.7} & \muted{67.8} & 53.1 & \muted{67.1} & \gapcell{+11.1} & +7.9
             & 56.7 & \muted{82.8} & \muted{60.2} & 42.8 & \muted{60.7} & \gapcell{+8.7}  & +4.2  \\
      & TCO  & 66.8 & \muted{90.8} & \muted{72.1} & 56.4 & \muted{70.5} & \gapcell{+15.4} & +11.2
             & 62.2 & \muted{90.0} & \muted{66.7} & 47.1 & \muted{65.9} & \gapcell{+14.2} & +8.5  \\

  \bottomrule
\end{tabular}%
}
\end{table*}
\endgroup

%% file: 4_experiments_tables_yt_offline.tex
\begingroup
\setlength{\tabcolsep}{2pt} 

\begin{table*}[t]
\centering
\caption{Def., TO, and TCO results for \textbf{offline} VIS methods on
YouTube-VIS 2019 and 2021~\cite{yang2019vis, yang2021vis}. TO isolates tracking error; TCO additionally
removes classification inconsistencies. $\Delta$AP and $\Delta$AR$_1$
are relative to the Default baseline.}
\label{tab:yt_offline}

% Use resizebox to force it to exactly the width of the page
\resizebox{\textwidth}{!}{%
\begin{tabular}{l l @{\hskip 2pt} ccccccc @{\hskip 4pt} ccccccc}
  \toprule
    \multirow{2}{*}{Method}
    & \multirow{2}{*}{Var.}
    & \multicolumn{7}{c}{YTVIS19}
    & \multicolumn{7}{c}{YTVIS21} \\
  \cmidrule(lr){3-9} \cmidrule(lr){10-16}
    &
    & AP & \muted{AP$_{50}$} & \muted{AP$_{75}$} & AR$_1$ & \muted{AR$_{10}$} & $\Delta$AP & $\Delta$AR$_1$
    & AP & \muted{AP$_{50}$} & \muted{AP$_{75}$} & AR$_1$ & \muted{AR$_{10}$} & $\Delta$AP & $\Delta$AR$_1$ \\
  \midrule

    \multirow{3}{*}{GenVIS~\cite{heo2023genvis}}
      & Def. & 48.4 & \muted{67.8} & \muted{54.3} & 45.0 & \muted{54.5} & --  & --
             & 46.4 & \muted{65.6} & \muted{51.6} & 38.4 & \muted{51.3} & --  & --  \\
      & TO   & 54.5 & \muted{76.3} & \muted{60.0} & 49.4 & \muted{63.1} & \gapcell{+6.1}  & +4.4
             & 50.4 & \muted{71.9} & \muted{54.9} & 42.8 & \muted{59.3} & \gapcell{+4.0}  & +4.5 \\
      & TCO  & 65.6 & \muted{91.8} & \muted{71.2} & 55.8 & \muted{70.5} & \gapcell{+17.2} & +10.8
             & 59.2 & \muted{83.6} & \muted{65.3} & 47.9 & \muted{65.1} & \gapcell{+12.9} & +9.5 \\

  \cmidrule(lr){1-16}

    \multirow{3}{*}{DVIS\cite{zhang2023dvis}}
      & Def. & 49.5 & \muted{71.8} & \muted{54.9} & 43.8 & \muted{55.3} & --  & --
             & 46.2 & \muted{69.3} & \muted{49.5} & 37.8 & \muted{53.3} & --  & --  \\
      & TO   & 50.1 & \muted{70.7} & \muted{54.1} & 48.4 & \muted{61.9} & \gapcell{+0.6}  & +4.6
             & 49.7 & \muted{71.0} & \muted{55.6} & 40.9 & \muted{59.0} & \gapcell{+3.5}  & +3.1 \\
      & TCO  & 66.0 & \muted{93.7} & \muted{70.2} & 55.5 & \muted{70.3} & \gapcell{+16.5} & +11.7
             & 61.0 & \muted{88.1} & \muted{67.7} & 46.3 & \muted{65.3} & \gapcell{+14.8} & +8.5 \\

  \cmidrule(lr){1-16}

    \multirow{3}{*}{DVIS++\cite{zhang2023dvisplus}}
      & Def. & 52.3 & \muted{75.6} & \muted{57.7} & 45.3 & \muted{57.7} & --  & --
             & 49.8 & \muted{72.4} & \muted{55.4} & 39.5 & \muted{56.0} & --  & --  \\
      & TO   & 60.0 & \muted{84.0} & \muted{66.6} & 51.9 & \muted{66.7} & \gapcell{+7.7}  & +6.6
             & 57.5 & \muted{80.2} & \muted{63.1} & 44.9 & \muted{64.3} & \gapcell{+7.7}  & +5.4 \\
      & TCO  & 68.7 & \muted{96.1} & \muted{76.6} & 57.3 & \muted{72.4} & \gapcell{+16.4} & +12.0
             & 65.4 & \muted{91.4} & \muted{71.5} & 48.6 & \muted{69.2} & \gapcell{+15.6} & +9.1 \\

  \cmidrule(lr){1-16}

    \multirow{3}{*}{CAVIS\cite{lee2024context}$^\dagger$}
      & Def. & 66.4 & \muted{87.2} & \muted{73.6} & 54.7 & \muted{71.0} & --  & --
             & 64.8 & \muted{86.5} & \muted{72.6} & 48.4 & \muted{69.3} & --  & --  \\
      & TO   & 73.8 & \muted{94.4} & \muted{82.4} & 59.7 & \muted{77.6} & \gapcell{+7.3}  & +4.9
             & 71.1 & \muted{94.0} & \muted{80.6} & 51.6 & \muted{75.7} & \gapcell{+6.3}  & +3.2 \\
      & TCO  & 77.9 & \muted{98.7} & \muted{87.8} & 63.0 & \muted{81.3} & \gapcell{+11.5} & +8.3
             & 72.3 & \muted{95.5} & \muted{82.5} & 52.4 & \muted{76.7} & \gapcell{+7.5}  & +4.1 \\

  \bottomrule
\end{tabular}%
}
\vspace{2pt}
{\footnotesize $^\dagger$All models are reported for R50 backbone except CAVIS, which uses ViT-L.}
\end{table*}
\endgroup

%% file: 4_experiments_tables_ovis.tex
\begin{table*}[htbp]
\centering
\caption{Def., TO, and TCO results for \textbf{online} VIS methods on the OVIS diagnostic split.
TO isolates tracking error by computing optimal query-to-instance assignments per frame.
$\Delta$AP and $\Delta$AR$_1$ are relative to the Default baseline.
Due to OVIS's extreme occlusion regime, gaps are substantially larger than on YouTube-VIS.}
\label{tab:ovis}

% \footnotesize
\resizebox{\textwidth}{!}{%

\setlength{\tabcolsep}{2pt}

\begin{tabular}{l l @{\hskip 2pt} ccccccc @{\hskip 4pt} ccccccc}
  \toprule
    \multirow{2}{*}{Method}
    & \multirow{2}{*}{Var.}
    & \multicolumn{7}{c}{ResNet-50}
    & \multicolumn{7}{c}{Swin-L} \\
  \cmidrule(lr){3-9} \cmidrule(lr){11-16}
    & & AP & \muted{AP$_{50}$} & \muted{AP$_{75}$} & AR$_1$ & \muted{AR$_{10}$} & $\Delta$AP & $\Delta$AR$_1$
      & AP & \muted{AP$_{50}$} & \muted{AP$_{75}$} & AR$_1$ & \muted{AR$_{10}$} & $\Delta$AP & $\Delta$AR$_1$ \\
  \midrule

    \multirow{3}{*}{MinVIS~\cite{huang2022minvis}}
      & Def. & 43.6 & \muted{63.7} & \muted{46.3} & 20.0 & \muted{46.4} & --  & --
             & 47.9 & \muted{67.6} & \muted{52.4} & 22.1 & \muted{51.2} & --  & -- \\
      & TO   & 66.4 & \muted{93.6} & \muted{74.1} & 23.8 & \muted{69.3} & \gapcell{+22.8} & +3.8
             & 71.8 & \muted{95.8} & \muted{81.8} & 24.7 & \muted{75.0} & \gapcell{+23.9} & +2.6 \\
      & TCO  & 67.6 & \muted{95.9} & \muted{74.7} & 24.3 & \muted{70.4} & \gapcell{+24.0} & +4.3
             & 73.6 & \muted{96.8} & \muted{86.0} & 25.2 & \muted{76.3} & \gapcell{+25.7} & +3.1 \\

  \cmidrule(lr){1-16}

  \multirow{3}{*}{VISAGE~\cite{kim2024visage}$^\dagger$}
      & Def. & 56.5 & \muted{77.5} & \muted{62.7} & 23.0 & \muted{58.0} & --  & --
             & \multicolumn{7}{c}{\muted{---}} \\
      & TO   & 72.2 & \muted{95.8} & \muted{83.0} & 25.1 & \muted{76.0} & \gapcell{+15.7} & +2.2
             & \multicolumn{7}{c}{\muted{---}} \\
      & TCO  & 72.6 & \muted{96.4} & \muted{83.6} & 25.4 & \muted{76.6} & \gapcell{+16.1} & +2.5
             & \multicolumn{7}{c}{\muted{---}} \\ 

  \cmidrule(lr){1-16}

    \multirow{3}{*}{CTVIS~\cite{ying2023ctvis}}
      & Def. & 47.3 & \muted{69.6} & \muted{50.1} & 20.0 & \muted{52.4} & --  & --
             & 50.2 & \muted{70.4} & \muted{53.9} & 21.8 & \muted{54.9} & --  & -- \\
      & TO   & 66.7 & \muted{94.3} & \muted{73.0} & 24.0 & \muted{69.1} & \gapcell{+19.4} & +4.0
             & 63.8 & \muted{86.2} & \muted{70.8} & 24.1 & \muted{70.5} & \gapcell{+13.6} & +2.3 \\
      & TCO  & 67.7 & \muted{96.3} & \muted{74.0} & 24.3 & \muted{69.9} & \gapcell{+20.4} & +4.3
             & 67.3 & \muted{91.5} & \muted{75.4} & 24.6 & \muted{71.8} & \gapcell{+17.1} & +2.8 \\

  \cmidrule(lr){1-16}

    \multirow{3}{*}{GenVIS~\cite{heo2023genvis}}
      & Def. & 49.7 & \muted{72.5} & \muted{50.6} & 21.2 & \muted{51.6} & --  & --
             & 54.6 & \muted{76.3} & \muted{59.0} & 22.9 & \muted{56.7} & --  & -- \\
      & TO   & 61.6 & \muted{90.9} & \muted{65.6} & 23.0 & \muted{65.9} & \gapcell{+11.9} & +1.8
             & 68.2 & \muted{92.2} & \muted{77.4} & 24.6 & \muted{70.9} & \gapcell{+13.6} & +1.7 \\
      & TCO  & 62.3 & \muted{92.0} & \muted{66.4} & 23.1 & \muted{66.4} & \gapcell{+12.6} & +2.0
             & 68.6 & \muted{92.8} & \muted{77.4} & 24.7 & \muted{71.0} & \gapcell{+14.0} & +1.8 \\

  \cmidrule(lr){1-16}

    \multirow{3}{*}{DVIS~\cite{zhang2023dvis}}
      & Def. & 48.6 & \muted{72.5} & \muted{51.3} & 21.0 & \muted{53.7} & --  & --
             & 56.9 & \muted{80.2} & \muted{62.7} & 22.2 & \muted{61.4} & --  & -- \\
      & TO   & 62.7 & \muted{91.7} & \muted{68.3} & 23.0 & \muted{65.7} & \gapcell{+14.1} & +2.0
             & 66.3 & \muted{90.0} & \muted{75.2} & 23.8 & \muted{70.0} & \gapcell{+9.4}  & +1.6 \\
      & TCO  & 63.0 & \muted{92.1} & \muted{68.6} & 23.5 & \muted{66.3} & \gapcell{+14.4} & +2.5
             & 68.2 & \muted{92.5} & \muted{77.8} & 24.3 & \muted{71.7} & \gapcell{+11.3} & +2.1 \\

  \cmidrule(lr){1-16}

    \multirow{3}{*}{DVIS++~\cite{zhang2023dvisplus}}
      & Def. & 52.4 & \muted{77.9} & \muted{55.3} & 22.1 & \muted{55.4} & --  & --
             & 60.7 & \muted{85.6} & \muted{66.7} & 23.3 & \muted{63.1} & --  & -- \\
      & TO   & 64.8 & \muted{93.7} & \muted{71.5} & 23.4 & \muted{67.4} & \gapcell{+12.4} & +1.3
             & 68.4 & \muted{92.0} & \muted{78.4} & 24.2 & \muted{71.7} & \gapcell{+7.7}  & +0.9 \\
      & TCO  & 64.0 & \muted{92.4} & \muted{70.3} & 23.6 & \muted{66.9} & \gapcell{+11.6} & +1.5
             & 69.2 & \muted{92.8} & \muted{79.5} & 24.4 & \muted{72.3} & \gapcell{+8.5}  & +1.1 \\

  \cmidrule(lr){1-16}

    \multirow{3}{*}{CAVIS~\cite{lee2024context}}
      & Def. & 54.6 & \muted{80.4} & \muted{59.5} & 22.3 & \muted{58.1} & --  & --
             & 59.2 & \muted{82.9} & \muted{65.6} & 23.0 & \muted{62.8} & --  & -- \\
      & TO   & 67.4 & \muted{95.1} & \muted{75.7} & 24.1 & \muted{69.9} & \gapcell{+12.8} & +1.8
             & 68.7 & \muted{92.5} & \muted{77.7} & 24.3 & \muted{71.9} & \gapcell{+9.5}  & +1.3 \\
      & TCO  & 67.8 & \muted{96.0} & \muted{76.3} & 24.3 & \muted{70.2} & \gapcell{+13.2} & +2.0
             & 69.1 & \muted{93.4} & \muted{78.3} & 24.5 & \muted{72.2} & \gapcell{+9.9}  & +1.5 \\

  \bottomrule
\end{tabular}%
}
\vspace{2pt}
{\footnotesize $^\dagger$Swin-L weights unavailable.}
\end{table*}

%% file: 4_experiments_plots_ytvis_scale.tex
% --- Okabe-Ito Academic Palette ---
\definecolor{r50base}{HTML}{0072B2}      % Strong Blue
\definecolor{swinbase}{HTML}{56B4E9}    % Sky Blue
\definecolor{trackgap}{HTML}{D55E00}     % Vermillion (Live Orange/Red)
\definecolor{classgap}{HTML}{009E73}        % Bluish Green
\definecolor{gridgray}{RGB}{230, 230, 230}

\begin{tikzpicture}

% =======================
% MAIN AXIS (LEFT, AP)
% =======================
\begin{axis}[
    name=main,
    ybar stacked,
    width=0.9\textwidth,
    height=0.375\textwidth,
    bar width=6pt,
    ymin=30, ymax=80, 
    xmin=0, xmax=42, 
    axis y line*=left,
    axis x line*=bottom,
    ylabel={Average Precision (AP)},
    ylabel style={font=\bfseries\scriptsize, yshift=-3pt},
    ytick={30, 40, 50, 60, 70, 80}, 
    yticklabel style={font=\bfseries\scriptsize},
    xtick={3, 10, 17, 24, 31, 38}, 
    xticklabels={MinVIS, CTVIS, GenVIS, DVIS, DVIS++, CAVIS}, 
    xticklabel style={
        rotate=45,
        anchor=north east,
        font=\bfseries\tiny,
        yshift=-15pt
    },
    legend style={
        at={(0.5, 1.10)},
        anchor=south,
        legend columns=-1,
        draw=none,
        font=\tiny,
        column sep=6pt,
        /tikz/every even column/.append style={column sep=6pt}
    },
    ymajorgrids=true,
    yminorgrids=true,
    minor y tick num=3, 
    major grid style={dashed, gridgray, line width=0.6pt},
    minor grid style={dotted, gridgray!60, line width=0.4pt},
    axis on top,
    clip=true 
]

% --- LAYER 1: BASE (Deep Blue) ---
\addplot[fill=r50base, draw=none] coordinates {
    % MinVIS
    (1, 42.9) (2, 0)   (4, 35.7) (5, 0)
    % CTVIS
    (8, 47.4) (9, 0)   (11, 37.3) (12, 0)
    % GenVIS
    (15, 46.3) (16, 0) (18, 39.2) (19, 0)
    % DVIS
    (22, 44.3) (23, 0) (25, 37.0) (26, 0)
    % DVIS++
    (29, 48.4) (30, 0) (32, 40.0) (33, 0)
    % CAVIS
    (36, 48.0) (37, 0) (39, 38.6) (40, 0)
};

% --- LAYER 2: SCALED (Light Blue) ---
\addplot[fill=swinbase, draw=none] coordinates {
    % MinVIS
    (1, 0) (2, 54.7)   (4, 0) (5, 44.8)
    % CTVIS
    (8, 0) (9, 60.1)   (11, 0) (12, 46.6)
    % GenVIS
    (15, 0) (16, 59.3) (18, 0) (19, 47.5)
    % DVIS
    (22, 0) (23, 57.9) (25, 0) (26, 45.7)
    % DVIS++
    (29, 0) (30, 61.3) (32, 0) (33, 47.8)
    % CAVIS
    (36, 0) (37, 59.4) (39, 0) (40, 46.0)
};

% --- LAYER 3: TRACKING GAP (Orange, Hatched) ---
\addplot[fill=trackgap, postaction={pattern=north east lines, pattern color=white!90}] coordinates {
    % MinVIS
    (1, 12.5) (2, 12.1)   (4, 7.9) (5, 5.8)
    % CTVIS
    (8, 14.4) (9, 11.2)   (11, 9.7) (12, 5.8)
    % GenVIS
    (15, 5.4) (16, 5.5) (18, 4.8) (19, 2.5)
    % DVIS
    (22, 7.1) (23, 7.5) (25, 3.9) (26, 2.3)
    % DVIS++
    (29, 9.5) (30, 10.6) (32, 4.8) (33, 3.2)
    % CAVIS
    (36, 8.6) (37, 5.6) (39, 4.2) (40, 1.8)
};

% --- LAYER 4: CLASSIFICATION GAP (Green, Crosshatch/Grid) ---
\addplot[fill=classgap, postaction={pattern=crosshatch, pattern color=white!80}] coordinates {
    % MinVIS
    (1, 7.5) (2, 4.1)   (4, 3.5) (5, 1.1)
    % CTVIS
    (8, 7.0) (9, 3.3)   (11, 3.2) (12, 0.9)
    % GenVIS
    (15, 9.2) (16, 2.9) (18, 3.9) (19, 1.1)
    % DVIS
    (22, 7.7) (23, 4.0) (25, 4.4) (26, 2.5)
    % DVIS++
    (29, 6.5) (30, 2.0) (32, 3.0) (33, 1.9)
    % CAVIS
    (36, 5.5) (37, 2.1) (39, 4.2) (40, 1.9)
};

\legend{Base Default, Scaled Default, Tracking Gap, Classification Gap}

% --- ANNOTATIONS (AP / AR1 Labels) ---
\pgfplotsinvokeforeach{0, 7, 14, 21, 28, 35}{
    \node [anchor=north, font=\sffamily\bfseries\scriptsize, color=r50base, yshift=-2pt]
        at (axis cs: #1 + 1.5, 30) {AP}; 
    \node [anchor=north, font=\sffamily\bfseries\scriptsize, color=black!60, yshift=-2pt]
        at (axis cs: #1 + 4.5, 30) {AR$_1$}; 
}

\end{axis}

% =======================
% SECOND AXIS (RIGHT, AR1)
% =======================
\begin{axis}[
    at={(main.south east)},
    anchor=south east,
    width=0.9\textwidth,
    height=0.375\textwidth,
    ymin=30, ymax=80, 
    xmin=0, xmax=42, 
    axis y line*=right,
    axis x line=none,
    ytick={30, 40, 50, 60, 70, 80}, 
    yticklabel style={font=\bfseries\scriptsize},
    ylabel={Average Recall (AR$_1$)},
    ylabel style={font=\bfseries\scriptsize, yshift=0pt},
    grid=none,
    tick align=outside,
    clip=true 
]

% --- TRACKING GAP LABELS ---
% MinVIS (Base AP)
\node[font=\scriptsize, anchor=center, inner sep=1pt, fill=white!90, draw=black!50, rounded corners=0.5pt]
    at (axis cs:1, 49.15) {12.5};
% MinVIS (Scaled AP)
\node[font=\scriptsize, anchor=center, inner sep=1pt, fill=white!90, draw=black!50, rounded corners=0.5pt]
    at (axis cs:2, 60.75) {12.1};
% MinVIS AR1 (Base AR1)
\node[font=\scriptsize, anchor=center, inner sep=1pt, fill=white!90, draw=black!50, rounded corners=0.5pt]
    at (axis cs:4, 39.65) {7.9};
% MinVIS AR1 (Scaled AR1)
\node[font=\scriptsize, anchor=center, inner sep=1pt, fill=white!90, draw=black!50, rounded corners=0.5pt]
    at (axis cs:5, 47.7) {5.8};

% CTVIS (Base AP)
\node[font=\scriptsize, anchor=center, inner sep=1pt, fill=white!90, draw=black!50, rounded corners=0.5pt]
    at (axis cs:8, 54.6) {14.4};
% CTVIS (Scaled AP)
\node[font=\scriptsize, anchor=center, inner sep=1pt, fill=white!90, draw=black!50, rounded corners=0.5pt]
    at (axis cs:9, 65.7) {11.2};
% CTVIS AR1 (Base AR1)
\node[font=\scriptsize, anchor=center, inner sep=1pt, fill=white!90, draw=black!50, rounded corners=0.5pt]
    at (axis cs:11, 42.15) {9.7};
% CTVIS AR1 (Scaled AR1)
\node[font=\scriptsize, anchor=center, inner sep=1pt, fill=white!90, draw=black!50, rounded corners=0.5pt]
    at (axis cs:12, 49.5) {5.8};

% GenVIS (Base AP)
\node[font=\scriptsize, anchor=center, inner sep=1pt, fill=white!90, draw=black!50, rounded corners=0.5pt]
    at (axis cs:15, 49.0) {5.4};
% GenVIS (Scaled AP)
\node[font=\scriptsize, anchor=center, inner sep=1pt, fill=white!90, draw=black!50, rounded corners=0.5pt]
    at (axis cs:16, 62.05) {5.5};
% GenVIS AR1 (Base AR1)
\node[font=\scriptsize, anchor=center, inner sep=1pt, fill=white!90, draw=black!50, rounded corners=0.5pt]
    at (axis cs:18, 41.6) {4.8};
% GenVIS AR1 (Scaled AR1)
\node[font=\scriptsize, anchor=center, inner sep=1pt, fill=white!90, draw=black!50, rounded corners=0.5pt]
    at (axis cs:19, 48.75) {2.5};

% DVIS (Base AP)
\node[font=\scriptsize, anchor=center, inner sep=1pt, fill=white!90, draw=black!50, rounded corners=0.5pt]
    at (axis cs:22, 47.85) {7.1};
% DVIS (Scaled AP)
\node[font=\scriptsize, anchor=center, inner sep=1pt, fill=white!90, draw=black!50, rounded corners=0.5pt]
    at (axis cs:23, 61.65) {7.5};
% DVIS AR1 (Base AR1)
\node[font=\scriptsize, anchor=center, inner sep=1pt, fill=white!90, draw=black!50, rounded corners=0.5pt]
    at (axis cs:25, 38.95) {3.9};
% DVIS AR1 (Scaled AR1)
\node[font=\scriptsize, anchor=center, inner sep=1pt, fill=white!90, draw=black!50, rounded corners=0.5pt]
    at (axis cs:26, 46.85) {2.3};

% DVIS++ (Base AP)
\node[font=\scriptsize, anchor=center, inner sep=1pt, fill=white!90, draw=black!50, rounded corners=0.5pt]
    at (axis cs:29, 53.15) {9.5};
% DVIS++ (Scaled AP)
\node[font=\scriptsize, anchor=center, inner sep=1pt, fill=white!90, draw=black!50, rounded corners=0.5pt]
    at (axis cs:30, 66.6) {10.6};
% DVIS++ AR1 (Base AR1)
\node[font=\scriptsize, anchor=center, inner sep=1pt, fill=white!90, draw=black!50, rounded corners=0.5pt]
    at (axis cs:32, 42.4) {4.8};
% DVIS++ AR1 (Scaled AR1)
\node[font=\scriptsize, anchor=center, inner sep=1pt, fill=white!90, draw=black!50, rounded corners=0.5pt]
    at (axis cs:33, 49.4) {3.2};

% CAVIS (Base AP)
\node[font=\scriptsize, anchor=center, inner sep=1pt, fill=white!90, draw=black!50, rounded corners=0.5pt]
    at (axis cs:36, 52.3) {8.6};
% CAVIS (Scaled AP)
\node[font=\scriptsize, anchor=center, inner sep=1pt, fill=white!90, draw=black!50, rounded corners=0.5pt]
    at (axis cs:37, 62.2) {5.6};
% CAVIS AR1 (Base AR1)
\node[font=\scriptsize, anchor=center, inner sep=1pt, fill=white!90, draw=black!50, rounded corners=0.5pt]
    at (axis cs:39, 40.7) {4.2};
% CAVIS AR1 (Scaled AR1)
\node[font=\scriptsize, anchor=center, inner sep=1pt, fill=white!90, draw=black!50, rounded corners=0.5pt]
    at (axis cs:40, 46.9) {1.8};

\end{axis}

\end{tikzpicture}

%% file: 4_experiments_plots_density.tex
% Colors — Okabe-Ito, consistent with Figure 2
\definecolor{colorMinVIS}{HTML}{0072B2}
\definecolor{colorCTVIS}{HTML}{D55E00}
\definecolor{colorGenVIS}{HTML}{E69F00}
\definecolor{colorDVIS}{HTML}{009E73}
\definecolor{colorVISAGE}{HTML}{CC79A7}
\definecolor{colorCAVIS}{HTML}{000000}

\begin{tikzpicture}[trim axis left, trim axis right]
\begin{axis}[
    width=\linewidth,
    height=5cm,
    ylabel={AP Tracking Gap},
    ylabel style={font=\small, yshift=-3pt},
    symbolic x coords={low,mid,high},
    xtick=data,
    xticklabels={Low\,(2--4), Mid\,(5--8), High\,(${\geq}9$)},
    xticklabel style={font=\scriptsize},
    y tick label style={font=\scriptsize,
                        /pgf/number format/fixed,
                        /pgf/number format/precision=0},
    enlarge x limits=0.25,
    ymin=0, ymax=40,
    ytick={0,10,20,30,40},
    ymajorgrids=true,
    yminorgrids=true,
    minor tick num=1,
    major grid style={dashed, gray!40, line width=0.5pt},
    minor grid style={solid,  gray!20, line width=0.4pt},
    % Export legend under name — placed manually in the figure wrapper
    legend to name=legendOracleGap,
    legend style={
        legend columns=6,
        font=\scriptsize,
        draw=black!30,
        fill=white,
        column sep=4pt,
    },
]

\addplot[color=colorMinVIS, mark=*, mark size=2pt, line width=1.2pt]
    coordinates {(low,16.67)(mid,27.04)(high,29.23)};
\addlegendentry{MinVIS}

\addplot[color=colorCTVIS, mark=square*, mark size=2pt, line width=1.2pt]
    coordinates {(low,12.34)(mid,17.90)(high,23.44)};
\addlegendentry{CTVIS}

\addplot[color=colorVISAGE, mark=triangle*, mark size=2.5pt, line width=1.2pt]
    coordinates {(low,10.89)(mid,23.93)(high,30.84)};
\addlegendentry{VISAGE}

\addplot[color=colorGenVIS, mark=diamond*, mark size=2.5pt, line width=1.2pt]
    coordinates {(low,5.42)(mid,14.88)(high,21.90)};
\addlegendentry{GenVIS}

\addplot[color=colorDVIS, mark=pentagon*, mark size=2.5pt, line width=1.2pt]
    coordinates {(low,10.16)(mid,14.64)(high,19.71)};
\addlegendentry{DVIS}

\addplot[color=colorCAVIS, mark=+, mark size=3pt, line width=1.2pt]
    coordinates {(low,6.35)(mid,15.57)(high,18.58)};
\addlegendentry{CAVIS}

\end{axis}
\end{tikzpicture}

%% file: 4_experiments_plots_length.tex
% Colors — Okabe-Ito, consistent with Figure 2
\definecolor{colorMinVIS}{HTML}{0072B2}
\definecolor{colorCTVIS}{HTML}{D55E00}
\definecolor{colorGenVIS}{HTML}{E69F00}
\definecolor{colorDVIS}{HTML}{009E73}
\definecolor{colorVISAGE}{HTML}{CC79A7}
\definecolor{colorCAVIS}{HTML}{000000}

\begin{tikzpicture}[trim axis left, trim axis right]
\begin{axis}[
    width=\linewidth,
    height=5cm,
    ylabel={AP Tracking Gap},
    ylabel style={font=\small, yshift=-3pt},
    symbolic x coords={short,medium,long,vlong},
    xtick=data,
    xticklabels={Short, Medium, Long, V.\,Long},
    xticklabel style={font=\scriptsize},
    y tick label style={font=\scriptsize,
                        /pgf/number format/fixed,
                        /pgf/number format/precision=0},
    enlarge x limits=0.12,
    ymin=0, ymax=50,
    ytick={0,10,20,30,40,50},
    ymajorgrids=true,
    yminorgrids=true,
    minor tick num=1,
    major grid style={dashed, gray!40, line width=0.5pt},
    minor grid style={solid,  gray!20, line width=0.4pt},
    % No legend here — shared legend is in density file
]

\addplot[color=colorMinVIS, mark=*, mark size=2pt, line width=1.2pt]
    coordinates {(short,18.84)(medium,19.16)(long,34.29)(vlong,40.11)};

\addplot[color=colorCTVIS, mark=square*, mark size=2pt, line width=1.2pt]
    coordinates {(short,12.34)(medium,15.59)(long,21.64)(vlong,30.51)};

\addplot[color=colorVISAGE, mark=triangle*, mark size=2.5pt, line width=1.2pt]
    coordinates {(short,13.99)(medium,18.69)(long,28.67)(vlong,27.78)};

\addplot[color=colorGenVIS, mark=diamond*, mark size=2.5pt, line width=1.2pt]
    coordinates {(short,7.25)(medium,11.30)(long,19.72)(vlong,21.21)};

\addplot[color=colorDVIS, mark=pentagon*, mark size=2.5pt, line width=1.2pt]
    coordinates {(short,11.78)(medium,10.80)(long,22.16)(vlong,22.69)};

\addplot[color=colorCAVIS, mark=+, mark size=3pt, line width=1.2pt]
    coordinates {(short,7.44)(medium,12.24)(long,15.53)(vlong,22.66)};

\end{axis}
\end{tikzpicture}

%% file: 5_discussion.tex
% \paragraph{Why this decomposition order}
% Our hierarchical oracle isolates tracking \emph{first}, then classification, and finally mask quality. 
% This sequence is not arbitrary, but reflects the fundamental nature of query-based VIS predictions.
% Each query produces a mask and a class distribution, but carries no inherent instance label. While per-frame IoU or class logits offer hints, they can conflict: a query may overlap moderately with multiple instances, or share a category with several targets, leaving assignment ambiguous.
% Attempting to isolate classification or mask errors \emph{before} resolving identity would force frame-local heuristics (e.g., greedy matching) that obscure whether a misprediction stems from poor temporal association or genuinely weak per-frame output.
% By solving tracking globally via ILP, we construct a definitive query-to-instance mapping across the entire video, eliminating ambiguity in subsequent classification and mask comparisons.

\paragraph{The offline evaluation paradox}
The standard VIS evaluation protocol is fundamentally offline, requiring complete spatiotemporal tracks with a single class label per instance. Even nominally ``online'' methods respect this by assigning classes via temporal logit averaging—a non-causal operation requiring the full video. Furthermore, volumetric IoU metrics heavily penalize partial tracklets: a single identity switch ruins the accumulated intersection while expanding the union, severely punishing models even if their per-frame masks are perfect. Because online methods must commit to tracking decisions without future context, this metric creates a compounding penalty that directly drives the length-dependent degradation observed in Fig.~\ref{fig:oracle_gap_combined}. Future benchmarks should explore decoupled or windowed evaluation regimes to more faithfully reflect the true capabilities of online designs.

\paragraph{Design insights}
Our findings crystallize several architectural lessons. First, integrating temporal context into the association phase—whether via explicit memory (GenVIS) or decoupled temporal refinement (DVIS, DVIS++)—yields significantly smaller tracking gaps than pure frame-to-frame similarity matching. Second, the DVIS offline results illustrate a cascading bottleneck effect: successfully resolving tracking (yielding a near-zero +0.6\,AP gap on YTVIS19) exposes classification as a massive, previously hidden residual error (+15.9\,AP). Third, while appearance-aware tracking (e.g., VISAGE, CAVIS) reduces fragility, it cannot fully overcome the structural limits of frame-local matching. Finally, backbone scaling halves classification gaps but leaves AP tracking gaps intact, proving these bottlenecks require orthogonal solutions: stronger representations for classification, and fundamentally rethought association logic for tracking.

\paragraph{Limitations and future directions}
Because OVIS validation labels are withheld, our diagnostic analysis relies on a stratified training subset. Furthermore, while our oracle quantifies \emph{how much} performance is lost and TrackLens exposes \emph{where} failures occur, neither prescribes \emph{how} to fix them. Promising directions to resolve these bottlenecks include learnable long-term memory, uncertainty-aware matching, and training objectives that explicitly penalize identity drift under occlusion.

%% file: 6_conclusion.tex
Video instance segmentation performance is the outcome of three coupled skills: recognition, segmentation, and temporal association. This paper introduced a model-agnostic ILP oracle that hierarchically decomposes the VIS error budget into tracking, classification, and mask-quality components.

Across modern VIS systems, our analysis reveals that temporal association is the critical bottleneck, particularly for online methods facing occlusion, dense scenes, and long videos. While classification acts as a secondary error source on standard benchmarks, it nearly vanishes under heavy occlusion once tracking is resolved, confirming that frame-level semantic recognition is surprisingly robust. Furthermore, backbone scaling improves object recall and shrinks classification gaps but leaves AP tracking deficits intact, proving that temporal fragility stems from algorithmic matching logic rather than representational limits. We also highlighted how standard volumetric metrics impose non-causal, compounding penalties on online methods, underscoring the need for decoupled evaluation regimes.

Finally, by pairing this quantitative oracle with TrackLens, an interactive query-level debugging tool, we translate tracking drift from a vague symptom into observable failure modes. Together, these tools provide a systematic foundation to shift VIS research from incremental metric gains toward the targeted design of robust, long-horizon association mechanisms.

%% file: 7_appendix.tex
\section{Effect of the Argmax Constraint}
\label{sec:app_argmax}

This appendix validates two methodological choices: using single-label assignment as an evaluation baseline, and enforcing class-consistency within the ILP tracking formulation.

\paragraph{Cost of single-label assignment}
To quantify the impact of the argmax constraint on default performance, we compare standard multi-label MinVIS inference against a single-label variant (where each track is assigned its highest-scoring class and duplicates are suppressed). As Table~\ref{tab:argmax_default} shows, the argmax constraint reduces AP by merely 1.5--1.9 points. This confirms that enforcing a single-label assumption imposes only a minor, bounded cost, making our reported baselines slightly conservative.

\input{7_appendix_tables_argmax_default}

\paragraph{Role of the class-consistency constraint}
We next verify that the ILP class-consistency constraint is actively necessary. We introduce T-Oracle\textsuperscript{free}, an unconstrained variant that maximizes total IoU regardless of class logits, then uses the model's predicted label for the resulting track. 

Table~\ref{tab:argmax_oracle} shows that removing the class constraint actually \emph{drops} oracle AP by 7.1--7.4 points. Without it, the ILP routinely assigns queries that geometrically overlap the target but possess conflicting class logits. Averaging these incoherent logits produces unreliable track-level predictions, severely penalizing AP. The constraint prevents this by ensuring tracks are both geometrically and semantically coherent, proving it is a fundamental requirement for establishing a meaningful tracking upper bound.

\input{7_appendix_tables_argmax_oracle}

\paragraph{ILP reliability}
Across MinVIS (ResNet-50), the solver reached its 180-second timeout without certifying optimality in fewer than 2\% of videos (1/302 YTVIS19, 2/421 YTVIS21, 2/100 OVIS). Complete fallback to default output occurred in 6/302 and 8/421 videos, mostly single-instance videos where no query assigns the ground-truth class as its argmax, making any class-consistent track impossible. Instance-level reduction (re-solving over N-1 instances) was needed in 14/302 and 11/421 videos. Both outcomes reflect the model's prediction limits rather than a flaw in the oracle.

\section{TrackLens Practicality}
\label{sec:app_tracklens}

\begin{figure}[hbt!]
    \centering
    \includegraphics[width=\linewidth]{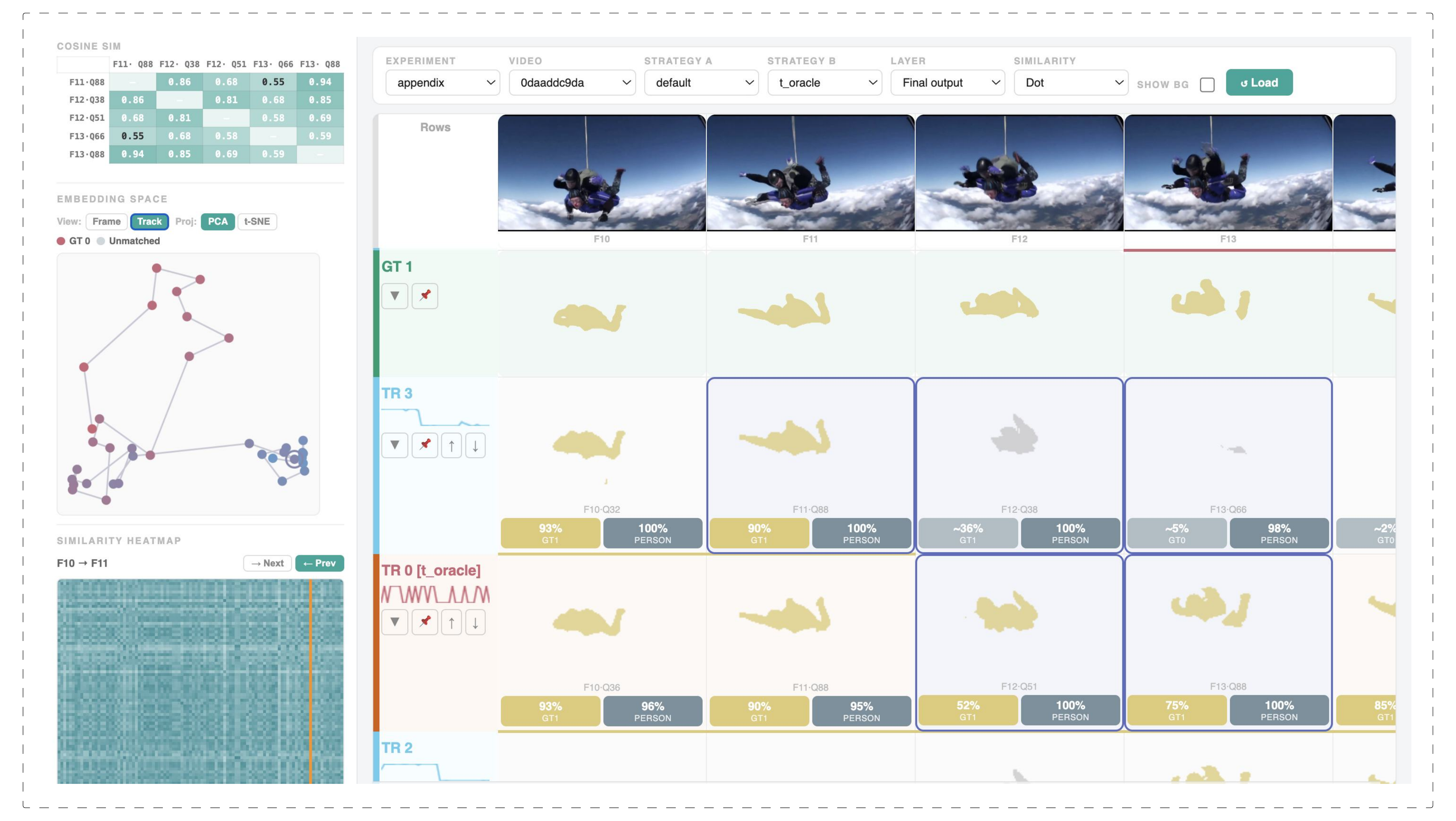}
    \caption{TrackLens visualization of a tracking failure in MinVIS on a
    representative video. The default track (blue row) incorrectly assigns Q38
    at frame 12, then propagates the error through frame 13 (Q66) and beyond.
    The T-Oracle track (orange row) shows the correct assignment sequence:
    Q52 at frame 12, Q88 at frame 13. Sidebar similarity matrices report
    cosine similarities between the frame-11 embedding and candidate queries
    at frame 13: 0.55 against the incorrectly assigned query and 0.94 against
    the correct one.}
    \label{fig:tracklens_practicality}
\end{figure}

TrackLens exposes failure modes by overlaying default and oracle tracking
strategies side-by-side and rendering per-query similarity matrices across
frames. Figure~\ref{fig:tracklens_practicality} shows a representative
identity switch: the default strategy assigns the wrong query at frame 12 and,
since each matching step conditions only on the immediately preceding frame,
cannot recover — propagating the error through subsequent frames. The sidebar
similarity matrices reveal why recovery was possible in principle but not
realized: the frame-11 embedding retains high similarity with the correct
frame-13 query (0.94) but low similarity with the one actually assigned (0.55).
Had earlier queries been consulted, the ambiguity would have resolved correctly.

This observation motivates a lightweight, training-free modification to the
MinVIS matching pipeline. At each frame $t$, MinVIS constructs a cosine
similarity matrix $S^{t,t-1}$ between the current queries and those of the
previous frame, then applies the Hungarian algorithm to obtain a query
assignment. We generalize this by replacing the single-frame matrix with a
temporally aggregated one,
\[
    S_{\text{agg}}^{t} = \sum_{i=1}^{W} w_i \cdot S^{t,\, t-i},
\]
where $S^{t,t-i}$ is the cosine similarity matrix between queries at frame $t$
and frame $t{-}i$, and weights $w_i$ sum to one. The Hungarian algorithm is
then applied to $S_{\text{agg}}^{t}$ identically to the original pipeline.

Table~\ref{tab:tracklens_window} shows that a window of $W{=}4$ yields a
consistent gain of approximately 3~AP on YTVIS19, in line with the finding
in VISAGE~\cite{kim2024visage} that memory-augmented matching improves tracking robustness.

\begin{table}[htb]
    \centering
    \caption{Effect of temporally aggregated similarity matching ($W{=}3$, uniform
    weights) on MinVIS, evaluated on YTVIS19.}
    \label{tab:tracklens_window}
    \begin{tabular}{lcccc}
        \toprule
        Method & AP & AP$_{50}$ & AR$_1$ & AR$_{10}$ \\
        \midrule
        MinVIS (default, $W{=}1$) & 45.3 & 66.7 & 42.5 & 50.6 \\
        MinVIS + aggregated matching & 47.9  & 69.9 & 44.4 & 53.15  \\
        \bottomrule
    \end{tabular}
\end{table}

%% file: 7_appendix_tables_argmax_default.tex
\begin{table}[htb]
\centering
\caption{Effect of single-label (argmax) enforcement on MinVIS default inference. $\Delta$ is relative to the multi-label variant.}
\label{tab:argmax_default}
\setlength{\tabcolsep}{5pt}
    \begin{tabular}{l l cc cc}
    \toprule
    Dataset & Variant & AP & AR$_1$ & $\Delta$AP & $\Delta$AR$_1$ \\
    \midrule
    \multirow{2}{*}{YTVIS19}
      & Multi-label  & 47.3 & 45.3 & --   & --   \\
      & Single-label & 45.3 & 42.6 & $-$1.9 & $-$2.7 \\
    \cmidrule(lr){1-6}
    \multirow{2}{*}{YTVIS21}
      & Multi-label  & 44.4 & 39.0 & --   & --   \\
      & Single-label & 42.9 & 35.7 & $-$1.5 & $-$3.4 \\
    \bottomrule
    \end{tabular}
\end{table}

%% file: 7_appendix_tables_argmax_oracle.tex
\begin{table}[htb]
    \centering
    \caption{Oracle AP for MinVIS with and without the class-consistency constraint. T-Oracle\textsuperscript{free} maximizes IoU without enforcing class correctness. T-Oracle enforces the constraint. Both retain the model's predicted label.}
    \label{tab:argmax_oracle}
    \setlength{\tabcolsep}{5pt}
    \begin{tabular}{l l cc}
    \toprule
    Dataset & Variant & AP & AR$_1$ \\
    \midrule
    \multirow{3}{*}{YTVIS19}
      & T-Oracle\textsuperscript{free} & 54.6 & 46.2 \\
      & T-Oracle                       & 62.0 & 52.6 \\
      & TC-Oracle                      & 69.8 & 57.8 \\
    \cmidrule(lr){1-4}
    \multirow{3}{*}{YTVIS21}
      & T-Oracle\textsuperscript{free} & 48.2 & 37.2 \\
      & T-Oracle                       & 55.4 & 43.6 \\
      & TC-Oracle                      & 62.9 & 47.1 \\
    \bottomrule
    \end{tabular}
\end{table}